\documentclass[10pt,twocolumn,letterpaper]{article}

\usepackage{caption}

\usepackage{iccv}
\usepackage{times}
\usepackage{epsfig}
\usepackage{graphicx}
\usepackage{amsmath}
\usepackage{amssymb}

\usepackage{graphicx}
\usepackage{amsmath}
\usepackage{bm}
\usepackage{amssymb}
\usepackage{booktabs}
\usepackage{enumerate}
\usepackage{enumitem}
\usepackage[accsupp]{axessibility}  

\newcommand{\E}{\mathbb{E}}
\newcommand{\Ss}{\mathcal{S}}

\usepackage{adjustbox}
\usepackage{multirow}
\usepackage{float}
\usepackage{array}
\newcolumntype{P}[1]{>{\centering\arraybackslash}p{#1}}
\newcolumntype{M}[1]{>{\centering\arraybackslash}m{#1}}

\usepackage{algorithm}
\usepackage{algorithmic}
\usepackage[most]{tcolorbox}
\tcbset{on line, 
        boxsep=4pt, left=0pt,right=0pt,top=0pt,bottom=0pt,
        colframe=white,
        highlight math style={enhanced}
        }

\usepackage{threeparttable}

\usepackage{listings}

\usepackage{courier}
\newcommand{\ct}[1]{\texttt{#1}}
\usepackage{physics}
\newcommand{\tta}{\vb*{\theta}}
\newcommand{\al}{\vb*{\alpha}}

\newcommand{\thickbar}[1]{\vb*{\bar{\text{$#1$}}}}
\newcommand{\thickTilde}[1]{\vb*{\Tilde{\text{$#1$}}}}

\usepackage[pagebackref=true,breaklinks=true,letterpaper=true,colorlinks,bookmarks=false]{hyperref}

\usepackage[capitalize]{cleveref}
\crefname{section}{Sec.}{Secs.}
\Crefname{section}{Section}{Sections}
\Crefname{table}{Table}{Tables}
\crefname{table}{Tab.}{Tabs.}

\iccvfinalcopy 

\def\iccvPaperID{7061} 

\ificcvfinal\fi

\begin{document}

\title{PGFed: Personalize Each Client's Global Objective for Federated Learning}

\author{Jun Luo$^{1}$
\quad
Matias Mendieta$^{2}$
\quad
Chen Chen$^{2}$
\quad
Shandong Wu$^{1,3,4,5}$ \\
$^{1}$ Intelligent Systems Program, University of Pittsburgh, Pittsburgh, PA\\
$^{2}$ Center for Research in Computer Vision, University of Central Florida, Orlando, FL\\
$^{3}$ Department of Radiology, University of Pittsburgh, Pittsburgh, PA\\
$^{4}$ Department of Biomedical Informatics, University of Pittsburgh, Pittsburgh, PA\\
$^{5}$ Department of Bioengineering, University of Pittsburgh, Pittsburgh, PA\\
{\tt\small jul117@pitt.edu \quad ma584394@ucf.edu \quad chen.chen@crcv.ucf.edu \quad wus3@upmc.edu}
}

\maketitle
\ificcvfinal\thispagestyle{empty}\fi

\begin{abstract}
   Personalized federated learning has received an upsurge of attention due to the mediocre performance of conventional federated learning (FL) over heterogeneous data. Unlike conventional FL which trains a single global consensus model, personalized FL allows different models for different clients. However, existing personalized FL algorithms only \textbf{implicitly} transfer the collaborative knowledge across the federation by embedding the knowledge into the aggregated model or regularization. We observed that this implicit knowledge transfer fails to maximize the potential of each client's empirical risk toward other clients. Based on our observation, in this work, we propose \textbf{P}ersonalized \textbf{G}lobal \textbf{Fed}erated Learning (\texttt{PGFed}), a novel personalized FL framework that enables each client to \textbf{personalize} its own \textbf{global} objective by \textbf{explicitly} and adaptively aggregating the empirical risks of itself and other clients. To avoid massive ($O(N^2)$) communication overhead and potential privacy leakage while achieving this, each client's risk is estimated through a first-order approximation for other clients' adaptive risk aggregation. On top of \ct{PGFed}, we develop a momentum upgrade, dubbed \ct{PGFedMo}, to more efficiently utilize clients' empirical risks. Our extensive experiments on four datasets under different federated settings show consistent improvements of \texttt{PGFed} over previous state-of-the-art methods. The code is publicly available at \url{https://github.com/ljaiverson/pgfed}.
\end{abstract}


\section{Introduction}
\label{sec:intro}

Recent years have witnessed the prosperity of federated learning (FL)~\cite{mcmahan2017communication,kairouz2019advances,li2020federateda,wang2021field} in collaborative machine learning where the participating clients are subject to strict privacy rules~\cite{voigt2017eu}. Conventional FL aims to train a single global consensus model by orchestrating the participating clients with a central server. The most notable FL algorithm, \ct{FedAvg}~\cite{mcmahan2017communication}, proceeds the training by exchanging between clients and server only the locally updated and globally aggregated model weights, leaving the private datasets intact. As a privacy-preserving machine learning technique, FL tremendously boosts new collaborations of decentralized parties in a number of areas~\cite{xu2021federated,khan2021federated,long2020federated,kairouz2019advances}.

Unfortunately, along with the emergence of conventional FL, new challenges have been posed in terms of systems and statistical heterogeneity~\cite{li2020federateda}. Systems heterogeneity addresses the variability of clients' computation abilities, sizes of storage, or even the choices of model architecture due to different hardware constraints of the clients. Whereas statistical heterogeneity, \textbf{the focus of this work}, refers to the non-IID data among the clients, which could lead to non-guaranteed convergence~\cite{li2020federated,sahu2018convergence} and poor generalizability performance~\cite{sattler2019robust}, even after fine-tuning~\cite{jiang2019improving}.


\begin{figure}[t]
  \centering
    \includegraphics[width=\linewidth]{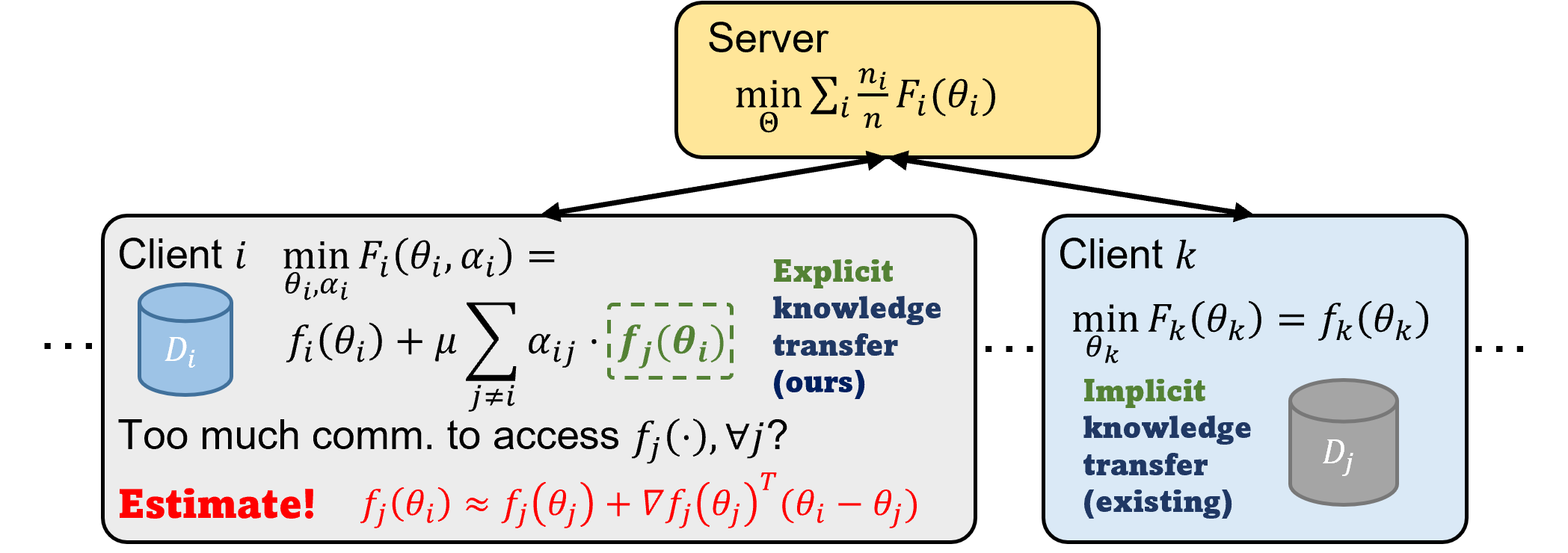}
    \caption{In \ct{PGFed}, the explicit collaborative knowledge transfer shows in the design of the local objective as a ``personalized global'' objective at client $i$, where non-local risks are involved.}
    \label{fig:PGFed}
\end{figure}

Over the years, the mediocre performance of conventional FL over heterogeneous data has not only promoted solutions that improve the single consensus model on top of \ct{FedAvg}~\cite{li2020federated,acar2020federated,karimireddy2020scaffold,li2021model}, a new paradigm, \textbf{\textit{personalized FL}}~\cite{tan2022towards,kulkarni2020survey}, has emerged as well. In this paradigm, personalized models are allowed for each individual client. Some efforts in this direction focus on different personalized layers and optimization techniques~\cite{liang2020think,collins2021exploiting,pillutla2022federated}. Leveraging multi-task~\cite{zhang2017survey} or meta-learning~\cite{hospedales2021meta} is also shown to be beneficial in personalized FL~\cite{smith2017federated,fallah2020personalized,chen2018federated}. Other works include clustered FL~\cite{ghosh2020efficient,shingi2020federated,sattler2020byzantine}, interpolation of personalized models~\cite{zhang2020personalized,luo2021adapt,deng2020adaptive}, and fine-tuning~\cite{oh2021fedbabu,li2019fedmd,wang2019federated}.

However, in most existing personalized FL algorithms~\cite{liang2020think,collins2021exploiting,deng2020adaptive,huang2021personalized,arivazhagan2019federated,fallah2020personalized,chen2022on,oh2021fedbabu}, \textit{the way in which the \textbf{collaborative knowledge} is transferred from the server to the clients is \textbf{implicit}.} Here, we consider the collaborative knowledge as non-local information, such as the global objective of \ct{FedAvg}, $F(\vb*{\theta})=\sum_i p_i F_i(\vb*{\theta})$, where $\vb*{\theta}$ is the global model and $F_i(\cdot)$ represents  client $i$'s local objective whose weights are denoted as $p_i$'s. In addition, we define \textit{``implicitness''} by defining its opposite side, \textit{``explicitness''}, as a direct engagement with multiple clients' empirical risks. For instance, updating the global model of \ct{FedAvg} is explicit, where the direct engagement is achieved through communication. However, this can hardly be the case for updating clients' personalized models, as it would take $O(N^2)$ communication overhead to transmit each client's personalized model to every client, assuming \ct{FedAvg}'s communication cost is $O(N)$ over $N$ clients. Consequently, most personalized FL algorithms implicitly transfer the collaborative knowledge from the server to the clients by embedding it into the aggregation of model weights or as different kinds of regularizers.



Why should we care about the ``\textit{explicitness}", especially for updating the personalized models? Let us first assume all communication has zero cost, and, as an example, design a client's explicit local objective as a \textit{``personalized global objective''} in the same form as the global objective of \ct{FedAvg} (weighted sum of all clients' risks), i.e. $F_i(\tta_i)=f_i(\tta_i)+\mu/(N-1) \sum_{j\neq i}f_j(\tta_i)$, where $F_i(\cdot)$ and $f_i(\cdot)$ are client $i$'s local objective and empirical risk, respectively, and $\mu$ is a hyperparameter. By this means, the clients are no longer limited to implicitly acquiring the collaborative knowledge in an embedded (from the aggregated model weights) form.

The motivation behind this explicit design is that it facilitates the generalizability of the personalized models by directly penalizing their performance over other clients' risks, contributing to more informative updates and therefore better local performance.
On the other hand, an implicit personalization scheme embeds the non-local information into the aggregated model weights, preventing the clients from directly engaging with other clients' risks.

We further demonstrate this motivation and the benefit of the explicit design by an empirical study: we personalize the output global model of \ct{FedAvg} through $S=6$ steps of local gradient-based update, supposing that the $O(SN^2)$ communication cost is affordable for now. We compare the exemplar explicit design of each client's local objective with a simple implicit method, i.e. the local fine-tuning of \ct{FedAvg} where each client's local objective is only its own empirical risk ($F_i(\tta_i)=f_i(\tta_i)$).

\begin{figure}[t] 
    \centering
    \includegraphics[width=0.98\linewidth]{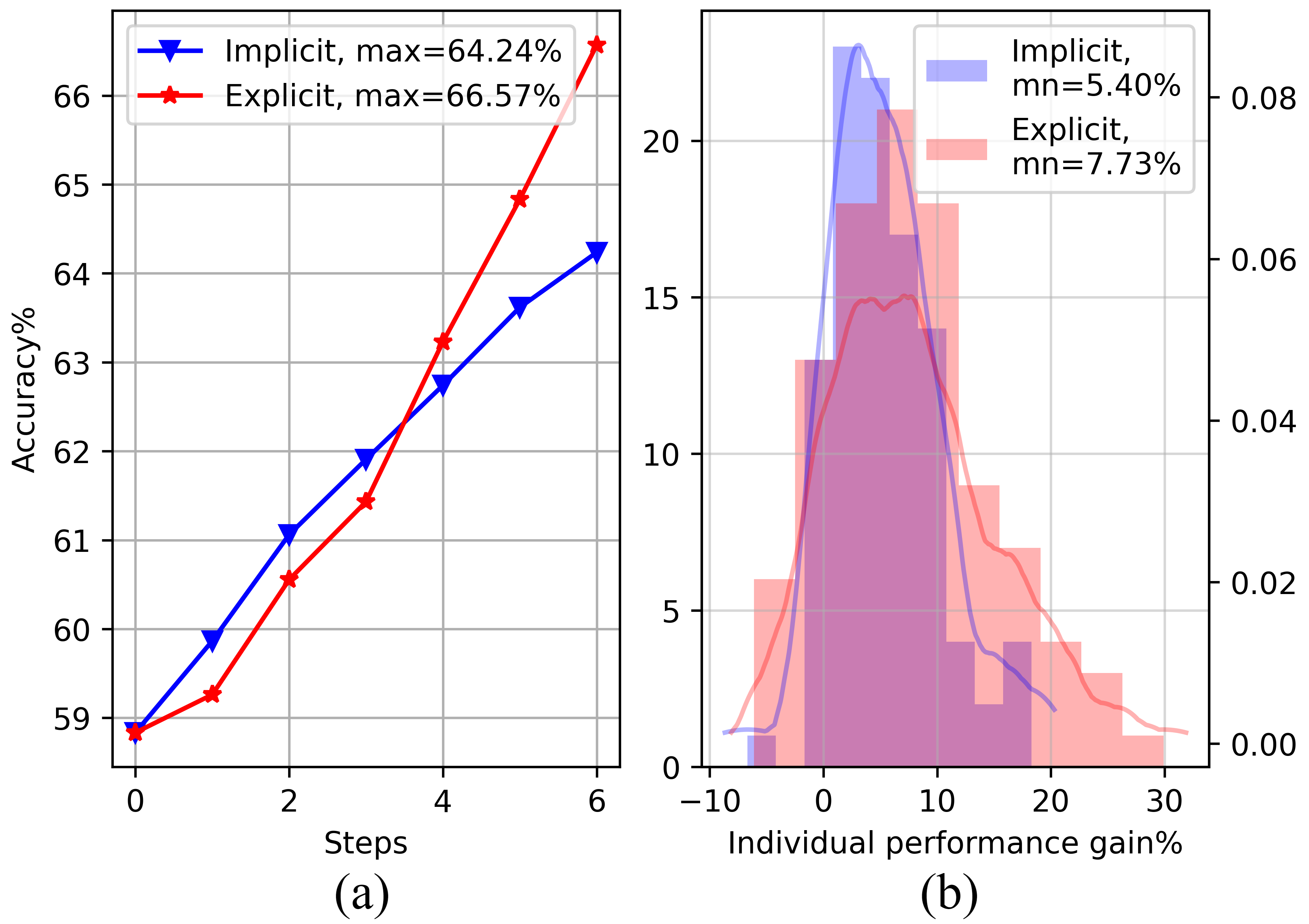}
    \label{fig:intro_expimp_a}
    \caption{In the task of personalizing the output of \ct{FedAvg} on CIFAR10 with 100 heterogeneous clients, the performance of the exemplar explicit local objective and an implicit local objective, assuming the communication cost is affordable. Figure (a) and (b) show the trend of the mean personalized test accuracy, and the histogram and density estimation of the individual gain, respectively.}
    \label{fig:intro_expimp}
\end{figure}

The result of this empirical study are shown in \cref{fig:intro_expimp}. On CIFAR10 dataset with 100 heterogeneous clients, we observe that the explicit transfer of collaborative knowledge exhibits a stronger ability to adapt the model towards clients' local data than the implicit counterpart, with a mean individual performance gain of 7.73\% over local test data from the initial global model,  2.33\% higher than that of an implicit personalization. \textit{Intriguing as the results may seem, how can we transfer this idea back to the communication-expensive real-world FL settings where acquiring all $f_j(\tta_i) \ \forall i,j \in [N]$ will cost $O(N^2)$ communication overhead?} Our solution is to estimate $f_j(\tta_i)$ by approximation as shown in \cref{fig:PGFed}.



Based on the above observation, in this work, we propose \textbf{P}ersonalized \textbf{G}lobal \textbf{Fed}erated Learning (\texttt{PGFed}), a novel personalized FL framework that enables each client to \textbf{personalize} its own \textbf{global} objective by \textit{explicitly and adaptively} aggregating the empirical risks of itself and other clients. To avoid massive communication overhead and potential privacy leakage, each client's risk is estimated through a first-order approximation for other clients' adaptive risk aggregation. Thereby, the clients are able to explicitly acquire the collaborative knowledge, and their personalized models can enjoy better generalizability. We summarize our contributions as follows:
\setlist{nolistsep}
\begin{itemize}[noitemsep,leftmargin=*] 
    \item We uncover that the explicitness of a personalized FL algorithm empowers itself with stronger adaptation ability. Based on this observation, we propose \ct{PGFed}, a novel explicit personalized FL algorithm that frames the local objective of each client as a personalized global objective.
    \item To the best of our knowledge, \ct{PGFed} is the first work in the field to achieve explicit transfer of global collaborative knowledge among the clients, without introducing the seemingly unavoidable $O(N^2)$ communication costs.
    \item On top of \ct{PGFed}, we develop a momentum upgrade, dubbed \ct{PGFedMo}, to let the clients more efficiently utilize other clients' empirical risks. 
    \item We evaluate \ct{PGFed} and \ct{PGFedMo} on four datasets under different FL settings. The results show that both algorithms outperform the compared state-of-the-art personalized FL methods, with up to \textbf{15.47\% boost in accuracy}.
\end{itemize}

\section{Related Work}
\label{sec:relatedworks}
\paragraph{Federated Learning.}
Federated learning~\cite{kairouz2019advances,li2020federateda,wang2021field} focuses on training a global consensus model over a federation of clients with similar data. 
The most notable FL algorithm, \ct{FedAvg}~\cite{mcmahan2017communication}, broadcast copies of the global model for local training and aggregate the updated copies as the new global model for the next round.

However, evidence began to accumulate in recent years that \ct{FedAvg} is vulnerable towards data heterogeneity (non-IID data)~\cite{li2020federated,sattler2019robust,sahu2018convergence,jiang2019improving}, which is the most common data distribution scenario in real-world FL. This vulnerability manifests itself as non-guaranteed convergence~\cite{sahu2018convergence} and poor generalizability on test data after fine-tuning~\cite{jiang2019improving}, which can cost the incentives of participating clients.

To improve the performance of the global model, \ct{FedProx}~\cite{li2020federated} adds a proximal term to clients' empirical risks to restrict the local update. \ct{FedDyn}~\cite{acar2020federated} introduces a dynamic regularizer to each client to align the global and local optima. \ct{SCAFFOLD}~\cite{karimireddy2020scaffold} corrects the local update by a control variate. And \ct{FedAlign}~\cite{mendieta2022local} focuses on local model generality rather than proximal restrictions. However, while effective, these global FL algorithms cannot systematically alleviate the data heterogeneity challenge.

\paragraph{Personalized Federated Learning.}
Personalized FL~\cite{tan2022towards,kulkarni2020survey} allows different models for different clients. This relaxation fundamentally mitigates the impact caused by the non-IID data. Methods in this category are the ones we mainly compare with in our experiments.

In this branch of work, \cite{arivazhagan2019federated,liang2020think,collins2021exploiting} focus on aggregating different layers of the model. A recent work, \ct{FedBABU}~\cite{oh2021fedbabu}, aggregates the feature extractor for a global model, and fine-tunes it for personalized models. Moreover, some works leverage multi-task~\cite{zhang2017survey} or meta-learning~\cite{hospedales2021meta} to learn the relationships between clients' data distributions~\cite{smith2017federated,fallah2020personalized,chen2018federated}. For instance, \ct{Per-FedAvg}~\cite{fallah2020personalized} takes advantage of Model-Agnostic Meta-Learning (MAML)~\cite{finn2017model} and trains an easy-to-adapt initial shared model. Instead of training a different model per client, clustered FL~\cite{ghosh2020efficient,shingi2020federated} trains a distinct model for each cluster of similar clients. Interpolation of personalized models is also a popular direction~\cite{zhang2020personalized,luo2021adapt,deng2020adaptive}, where works concentrate more on a personalized way to aggregate the models from different clients.

In addition, some recent works~\cite{zhang2020personalized,luo2021adapt} afford to pay the $O(N^2)$ communication cost to transmit the clients' models to other clients, but their massive communication cost only benefits the model aggregation, instead of paying more attention on aggregating clients' risks, which makes their algorithms fall into the implicit category. Another recent work~\cite{chen2022on} bridges the personalized FL and global FL by training a global model with balanced risk and personalized adaptive predictors with empirical risk. However, it is still implicit, as the clients do not engage with others' risks. In our work, we focus on an explicit way of personalization by introducing estimates of non-local empirical risks to each client without massive communication costs.

\section{Problem Formulation}
\label{sec:problemformulation}
In this section, we formalize the problem of conventional and personalized FL, and further elaborate on the explicitness of personalized FL.

\subsection{Conventional and Personalized FL}
Conventional federated learning aims to train a global consensus model for a federation of clients with similar data. Take the most notable FL algorithm, \ct{FedAvg}~\cite{mcmahan2017communication}, as an example. For a federation of $N$ clients, the goal of \ct{FedAvg} is to minimize the global objective defined as:
\begin{equation}
    \min_{\tta} F(\tta)= \sum_{i=1}^N p_i F_i(\tta),
    \label{eq:global_obj}
\end{equation}
where $\tta$ denotes the global model, $F_i(\cdot)$ represents the local objective of client $i$, and the weight $p_i$ is often set as $p_i=n_i/n$ with $n=\sum_k n_k$ where $n_k$ denotes the number of data samples on client $k$. In \ct{FedAvg}, the local objective $F_i(\cdot)$ measures client $i$'s empirical risk, i.e.
\begin{equation}
    F_i(\tta)=\E_{\xi \sim \mathcal{D}_{i}} f_i(\tta |\xi) \approx \sum_{k=1}^{n_i} f_i(\tta |\xi_k),
    \label{eq:local_imp}
\end{equation}
where $\mathcal{D}_i$ represents the data distribution, $\xi_k$ is the $k$-th data sample on client $i$, and the empirical risk $f_i(\cdot)$ is used to approximate the true risk on client $i$. To simplify the notations, in the rest of the paper, we drop the summation, and denote $f_i(\cdot)$ directly as $\sum_{k=1}^{n_i} f_i(\cdot |\xi_k)$ unless further clarified.

Personalized FL relaxes the number of models. Compared to conventional FL, each client $i$ is allowed to have its own personalized model $\tta_i$, i.e. the goal of personalized FL is defined as:
\begin{equation}
    \min_{\vb*{\Theta}} F(\vb*{\Theta})=\min_{\tta_1,...,\tta_N}\sum_{i=1}^N p_i F_i(\tta_i),
    \label{eq:pfl_global_obj}
\end{equation}
where $\vb*{\Theta}$ is a $d\times N$ matrix with $d$ being the number of dimensions of the model.

In addition, the number of participating clients ($N$) in FL is often large~\cite{kairouz2019advances}, and not all clients are able to participate in each federated round. Therefore, a subset of clients $\Ss$ with $|\Ss|=M$ is selected for each round of training.

\subsection{Implicit vs. Explicit Local Objective}
\label{sec:implicit_vs_explicit}
As discussed in \cref{sec:intro}, we defined two types of ways for transferring the collaborative knowledge from the server to the clients for personalized FL algorithms. The explicit way updates the model with a direct engagement with multiple clients' empirical risks. For instance, updating the global model of \ct{FedAvg} achieves explicitness through communication. And in the empirical study (\cref{fig:intro_expimp}) in \cref{sec:intro}, we provided an example of explicit local objective as:

\begin{equation}
    F_i(\tta_i)=f_i(\tta_i)+\frac{\mu}{(N-1)} \sum_{j\neq i}f_j(\tta_i).
    \label{eq:exp_exemplar}
\end{equation}

It is obvious that engaging all $f_j(\tta_i) \ \forall i,j \in [N]$ to update the clients' personalized models would take $O(N^2)$ communication overhead. Therefore, most personalized FL algorithms implicitly transfer the collaborative knowledge by embedding it into the aggregation of model weights or as different regularizers.

One pitfall is that, to see whether a personalized FL algorithm manages to afford the $O(N^2)$ communication overhead is \textit{NOT} the criterion of checking its explicitness. ~\cite{zhang2020personalized,luo2021adapt} are two recent works that did afford this communication cost on small federations, but since their personalized model updates do not directly involve other clients' risks, both of these algorithms still fall into the implicit category.

In addition, note that the explicitness of a personalized FL algorithm does not prevent itself from simultaneously possessing characteristics from the implicit counterpart. For instance, in a personalized FL algorithm, taking the globally aggregated model as a round-beginning initialization for clients' personalized models is an implicit characteristic that can co-exist with using \cref{eq:exp_exemplar} as an explicit personalization.

\section{Method}
\label{sec:method}
In this section, we introduce the proposed algorithm, \ct{PGFed}, in details. We clarify our global and local objectives, explain how we circumvent the massive communication cost, and introduce a momentum version of the algorithm, dubbed \ct{PGFedMo}. The full procedure of \ct{PGFed} is provided in \cref{alg:PGFed}.

\subsection{Objectives of \textbf{\ct{PGFed}}}
\label{sec:objs_of_pgfed}
We adopt \cref{eq:pfl_global_obj} as the global objective of \ct{PGFed}. As mentioned in \cref{sec:implicit_vs_explicit}, an algorithm can simultaneously leverage characteristics from both explicit and implicit collaborative knowledge transfer. In each training round, we first implicitly transfer the collaborative knowledge to the clients by embedding it into the globally aggregated model, $\tta_{glob}$, which serves as the round-beginning initialization for the clients' personalized models.

To address the explicitness, we design the local objective as a \textit{personalized global objective} in the same form as the global objective of \ct{FedAvg} (weighted sum of every clients' risks). Different from the previous exemplar design shown in \cref{eq:exp_exemplar}, in \ct{PGFed}, the local objective is defined as:
\begin{equation}
    F_i(\tta_i, \al_i)=f_i(\tta_i)+ \mu \sum_{j\in [N]} \alpha_{ij}f_j(\tta_i),
    \label{eq:pgfed_local_obj}
\end{equation}
where $\al_i$ is a $N$-dimensional vector of \textbf{learnable} scalars $\alpha_{ij}>0$ which denotes the personalized weights of $f_j(\cdot)$ on client $i$. In this way, each client is able to personalize how much other clients' risks should contribute to their own objective. Since $M$ clients are selected for every round, we initialize every as $\alpha_{ij}=1/M \ \forall i, j\in [N]$. This new local objective slightly changes the global objective to:
\begin{equation}
    \min_{\vb*{\Theta}, \vb*{A}} F(\vb*{\Theta}, \vb*{A})=\min_{\tta_1,...,\tta_N,\al_1,...,\al_N}\sum_{i=1}^N p_i F_i(\tta_i, \al_i),
    \label{eq:pgfed_local_per_obj}
\end{equation}
where $\vb*{A}$ is an $N\times N$ matrix with the $i$-th row being $\al_i$.

Now, the question becomes: how can we overcome the massive communication cost to achieve explicit personalization? We answer this question in the next subsection.

\subsection{Non-local Risk Estimation}
In most real-world FL settings, $O(N^2)$ communication cost is often not affordable, and \ct{PGFed} is no exception. To achieve the explicitness without having to pay the unaffordable cost, we estimate the non-local empirical risks by first-order approximations. Specifically, we define the non-local empirical risk on client $i$ as $f_j(\tta_i) \ \forall j\neq i$, and estimate this term by using Taylor expansion at $\tta_j$, i.e.
\begin{equation}
    f_j(\tta_i) \approx f_j(\tta_j) + \nabla_{\tta_j}f_j(\tta_j)^T(\tta_i-\tta_j),
    \label{eq:taylor_exp}
\end{equation}
where the higher-order terms are ignored. By plugging the approximation into \cref{eq:pgfed_local_obj}, we have:
\begin{equation}
    F_i(\tta_i, \al_i)\approx f_i(\tta_i)+ \mathcal{R}_{aux}^{[N]}(\tta_i, \al_i),
    \label{eq:pgfed_taylor_obj}
\end{equation}
\begin{equation}
\resizebox{1.0\hsize}{!}{$\mathcal{R}_{aux}^{[N]}(\tta_i, \al_i) = \mu \sum_{j\in [N]} \alpha_{ij} \left( f_j(\tta_j) + \nabla_{\tta_j}f_j(\tta_j)^T(\tta_i-\tta_j) \right).$}
    \label{eq:r_aug}
\end{equation}
where we define $\mathcal{R}_{aux}^{[N]}(\cdot)$ as an \textit{auxiliary risk} over all client $j\in[N]$. It is the existence of the auxiliary risk $\mathcal{R}_{aux}^{[N]}(\cdot)$ that makes the proposed algorithm an explicit personalization.

The intuition behind why the approximation might work lies in the regularization effect of our explicit personalization. As we personalize different clients' models in the same form of global objective (\cref{eq:pgfed_local_obj}), the non-local risks restrain the personalized model weights from ungoverned drifting. Compared to an implicit setting (e.g. \ct{FedAvg}~\cite{mcmahan2017communication}), the explicitly personalized models are more regularized, which enables the clients to train their personalized models in a more uniform direction. Therefore, as the regularization effect reduces the gaps between the personalized models, the first-order approximation is rewarded with a chance to shine in the proposed explicit personalization.

\begin{algorithm}[tb]
    \caption{\ct{PGFed} and \ct{PGFedMo}}
    \label{alg:PGFed}
    \textbf{Input:} $N$ clients, learning rates $\eta_1, \eta_2$, number of rounds $T$, coefficient $\mu$(, momentum $\beta$ for \ct{PGFedMo})\\
    \textbf{Output:} Personalized models $\tta_1^T,...,\tta_N^T$.\\
    \tcbox[colback=blue!20!white]{
    \begin{minipage}{1\linewidth}
    \textbf{ServerExecute:}
        \begin{algorithmic}[1] 
            \STATE Initialize $\alpha_{ij}\leftarrow 1/M \ \forall i,j\in [N]$, global model $\tta_{glob}^{0}$
            \STATE $\vb{A}[i] \leftarrow \al_i \ \forall i\in[N]$
            \FOR{$t \leftarrow 1,2,...,T$}
                \STATE Select a subset of $M$ clients, $\Ss_t$
                \STATE $g^{(1)}_t\leftarrow\{\}$; $\nabla_t\leftarrow\{\}$ \textcolor{blue}{// built for next round}
                \FOR{$i \in \Ss_t$ \textbf{in parallel}}
                    \IF{t=1}
                        \STATE $\tta_i^t, g_{\alpha}^{(1)}, \nabla f(\tta_{i}^{t}), \al_i\leftarrow$ \textbf{ClientUpdate}($\tta_{glob}^{t-1}$, $t$)
                    \ELSE
                        \STATE $\thickTilde{g}_{\Ss_{t-1}}\leftarrow \mu \sum_{j\in \Ss_{t-1}} \alpha_{ij} \nabla_{t-1}[j]$
                        \STATE $\thickbar{g}_{\Ss_{t-1}}\leftarrow\frac{\mu}{M} \sum_{j\in \Ss_{t-1}}\nabla_{t-1}[j] $
                        \STATE $\tta_i^t, g_{\alpha}^{(1)}, \nabla f(\tta_{i}^{t}), \al_i\leftarrow$ \textbf{ClientUpdate}($\tta_{glob}^{t-1}$, $t$, $\thickTilde{g}_{\Ss_{t-1}}$, $\thickbar{g}_{\Ss_{t-1}}$, $g^{(1)}_{t-1}$)
                    \ENDIF
                    \STATE \textcolor{blue}{// the next line records the values for next round}
                    \STATE $\vb{A}[i]\leftarrow \al_i$; $g^{(1)}_t[i]\leftarrow g_{\alpha}^{(1)}$; $\nabla_t[i] \leftarrow \nabla f(\tta_{i}^{t})$
                    \STATE \label{alg:globmodel}$\tta_{glob}^{t} \leftarrow \sum_{i\in \Ss_t}p_i \tta_i^t$
                    
                \ENDFOR
                \FOR{$i \in ([N]-\Ss_t)$ \textbf{in parallel}}
                    \STATE $\tta_i^t\leftarrow\tta_i^{t-1}$; $\thickTilde{g}_i^t\leftarrow \thickTilde{g}_i^{t-1}$
                \ENDFOR
            \ENDFOR
            \STATE \textbf{return} $\tta_1^T,...,\tta_N^T$
        \end{algorithmic}
    \end{minipage}}
    \tcbox[colback=red!30!white]{
    \begin{minipage}{1\linewidth}
    \textbf{ClientUpdate}($\tta_{global}^{t-1}$, $t$ (, $\thickTilde{g}$, $\thickbar{g}$, $g^{(1)}_{t-1}$)):\\
        \vspace{-1em}
        \begin{algorithmic}[1]
        \IF{t=1}
            \STATE $\tta_{i}^{t} \leftarrow$ \textbf{ClientUpdate}($\tta_{global}^{t-1}, \eta_1$) as in \ct{FedAvg}
        \ELSE
            \STATE $\tta_{i}^{t} \leftarrow \tta_{global}^{t-1}$
            \STATE $\thickTilde{g}_i^t \leftarrow \thickTilde{g}$ \textcolor{blue}{// without momentum}
            \STATE $\thickTilde{g}_i^t \leftarrow (1-\beta) \thickTilde{g} + \beta\thickTilde{g}_i^{t-1}$ \textcolor{blue}{// with momentum}
            \FOR{Batch of data $\mathcal{B}\in\mathcal{D}_i$}
                \STATE $\tta_{i}^{t} \leftarrow \tta_{i}^{t} - \eta_1(\nabla f(\tta_{i}^{t}, \mathcal{B})+\thickTilde{g}_t^i)$
                \STATE $g^{(2)}=\thickbar{g}^T\tta_i$
                \STATE $\forall j\in g^{(1)}_{t-1}: \ \alpha_{ij} \leftarrow \alpha_{ij} - \eta_2 (g^{(1)}_{t-1}[j] + g^{(2)})$
            \ENDFOR
        \ENDIF
        \STATE $g_{\alpha}^{(1)}\leftarrow\mu \left( f(\tta_{i}^{t}) - \nabla f(\tta_{i}^{t})^T\tta_{i}^{t} \right)$ \textcolor{blue}{// for next round}
        \STATE \textbf{return} $\tta_{i}^{t}, g_{\alpha}^{(1)}, \nabla f(\tta_{i}^{t}), \al_i$
        \end{algorithmic}
    \end{minipage}}
\end{algorithm}

\subsection{Gradient-based Update}
In \ct{PGFed}, the personalized models are updated through gradient-based optimizers such as stochastic gradient descent (SGD). In this section, we derive the gradient of the local objective of client $i$ with respect to the personalized model $\tta_i$ and the personalized weights $\al_i$ for the non-local risks. Note that the gradient of the local objective is just the gradient of the local empirical risk plus the gradient of the auxiliary risk. Therefore, for the gradient w.r.t. $\tta_i$, we have:
\begin{equation}
    \begin{split}
        \nabla_{\tta_i}F_i(& \tta_i, \al_i) = \nabla_{\tta_i}f_i(\tta_i) + \nabla_{\tta_i} \mathcal{R}_{aux}^{[N]}(\tta_i, \al_i)\\
        & = \nabla_{\tta_i}f_i(\tta_i) + \underbrace{\mu \sum_{j\in [N]} \alpha_{ij} \nabla_{\tta_j}f_j(\tta_j)}_\text{$\thickTilde{g}_{[N]}$}.
    \end{split}
    \label{ep:grad_theta_i}
\end{equation}
Since the $\thickTilde{g}_{[N]}$, the \textit{auxiliary gradient}, is not related to $\tta_i$, we can have it computed by the server by acquiring $\al_i$ from client $i$, and $\nabla_{\tta_j}f_j(\tta_j)$ from client $j$.

\begin{table*}[t]
    \centering
    \begin{tabular}{wl{3.0cm}|M{1.8cm}M{1.8cm}M{1.8cm}|M{1.8cm}M{1.8cm}M{1.8cm}}
    \toprule
     & \multicolumn{3}{c|}{CIFAR10} & \multicolumn{3}{c}{CIFAR100} \\ 
     & 25 clients & 50 clients & 100 clients & 25 clients & 50 clients & 100 clients \\ \midrule
    \ct{Local} & $72.40\pm0.45$ & $70.28\pm0.38$ & $67.39\pm0.20$ & $32.74\pm0.08$ & $26.05\pm0.34$ & $23.06\pm0.47$ \\
    \ct{FedAvg}~\cite{mcmahan2017communication} & $65.07\pm0.25$ & $64.41\pm0.66$ & $63.19\pm0.46$ & $28.48\pm0.59$ & $26.06\pm0.65$ & $25.58\pm0.80$ \\
    \ct{FedDyn}~\cite{acar2020federated} & $67.31\pm0.36$ & $65.02\pm0.91$ & $62.49\pm0.06$ & $34.17\pm0.43$ & $27.06\pm0.18$ & $23.88\pm0.36$ \\ \midrule
    \ct{pFedMe}~\cite{t2020personalized} & $70.60\pm0.23$ & $68.92\pm0.35$ & $66.40\pm0.04$ & $27.97\pm0.24$ & $23.82\pm0.06$ & $22.35\pm0.03$ \\
    \ct{FedFomo}~\cite{zhang2020personalized} & $72.33\pm0.03$ & $72.17\pm0.48$ & $70.86\pm0.27$ & $32.15\pm0.61$ & $25.90\pm1.17$ & $24.48\pm0.44$ \\
    \ct{APFL}~\cite{deng2020adaptive} & $77.03\pm0.26$ & $77.36\pm0.18$ & $76.29\pm0.13$ & $39.16\pm0.93$ & $35.15\pm0.65$ & $33.86\pm0.60$ \\
    \ct{FedRep}~\cite{collins2021exploiting} & $76.85\pm0.44$ & $76.03\pm0.17$ & $72.30\pm0.52$ & $33.43\pm0.80$ & $26.86\pm0.39$ & $22.76\pm0.45$ \\
    \ct{LG-FedAvg}~\cite{liang2020think} & $72.83\pm0.28$ & $70.44\pm0.31$ & $67.55\pm0.09$ & $33.65\pm0.19$ & $27.13\pm0.37$ & $24.82\pm0.28$ \\
    \ct{FedPer}~\cite{arivazhagan2019federated} & $77.84\pm0.18$ & $77.76\pm0.22$ & $75.01\pm0.20$ & $35.22\pm0.67$ & $28.63\pm0.70$ & $25.56\pm0.26$ \\
    \ct{Per-FedAvg}~\cite{fallah2020personalized} & $75.49\pm0.74$ & $76.27\pm0.50$ & $75.41\pm0.35$ & $32.89\pm0.43$ & $32.24\pm0.75$ & $32.59\pm0.21$ \\
    \ct{FedRoD}~\cite{chen2022on} & $79.73\pm0.68$ & $79.61\pm0.22$ & $77.76\pm0.32$ & $39.55\pm0.58$ & $33.87\pm2.42$ & $31.49\pm0.19$ \\
    \ct{FedBABU}~\cite{oh2021fedbabu} & $78.92\pm0.36$ & $79.35\pm0.84$ & $76.34\pm0.22$ & $32.71\pm0.23$ & $29.66\pm0.64$ & $27.72\pm0.11$ \\ \midrule
    \textbf{\ct{PGFed} (ours)} & \underline{$81.02\pm0.41$} & \underline{$81.42\pm0.31$} & \underline{$78.56\pm0.35$} & \underline{$43.12\pm0.03$} & \underline{$38.45\pm0.44$} & \underline{$35.71\pm0.54$} \\
    \textbf{\ct{PGFedMo} (ours)} & $\vb{81.20}$$\pm$$\vb{0.08}$ & $\vb{81.48}$$\pm$$\vb{0.32}$ & $\vb{78.74}$$\pm$$\vb{0.22}$ & $\vb{43.44}$$\pm$$\vb{0.14}$ & $\vb{38.50}$$\pm$$\vb{0.45}$ & $\vb{35.76}$$\pm$$\vb{0.65}$\\
    \bottomrule
    \end{tabular}
    \caption{Mean top-1 personalized accuracy of the proposed algorithms and the baselines. We report the mean and standard deviation over three different seeds. The highest and second-highest accuracies under each setting are in \textbf{bold} and \underline{underlined}, respectively. \textbf{Within the comparison of personalized FL algorithms, \ct{PGFed} and \ct{PGFedMo} boost the accuracy by up to 15.47\%.}}
    \label{tab:main_results}
\end{table*}

For the gradient w.r.t. $\al_i$, similarly, we have:
\begin{equation}
    \begin{split}
        & \nabla_{\alpha_{ij}}F_i(\tta_i, \al_i) = \mu \left( f_j(\tta_j) + \nabla_{\tta_j}f_j(\tta_j)^T(\tta_i-\tta_j) \right)\\
        & = \underbrace{\mu \left( f_j(\tta_j) - \nabla_{\tta_j}f_j(\tta_j)^T\tta_j \right)}_\text{$g_{\alpha}^{(1)}$} + \underbrace{\mu \nabla_{\tta_j}f_j(\tta_j)^T\tta_i}_\text{$g_{\alpha}^{(2)}$}.
    \end{split}
    \label{ep:grad_alpha_ij}
\end{equation}
As shown in \cref{ep:grad_alpha_ij}, this gradient can be split into two components: the first term, $g_{\alpha}^{(1)}$, is a scalar purely associated with client $j$, which can be uploaded from client $j$ with little cost; the second term is an interaction between the gradient of client $j$ and the personalized model of client $i$. To accurately acquire $g_{\alpha}^{(2)}$ term itself would, again, cost $O(N^2)$ communication overhead. Therefore, we approximate this term by an average, $\thickbar{g}_{[N]}$, computed by the server, i.e.
\begin{equation}
    g_{\alpha}^{(2)} \approx \thickbar{g}_{[N]}^T\tta_i= \frac{\mu}{N} \left(\sum_{j\in [N]}\nabla_{\tta_j}f_j(\tta_j) \right)^T \tta_i.
    \label{ep:grad_alpha_2_mean}
\end{equation}
Although transmitting $\thickbar{g}_{[N]}$ can be costly, and it is possible for the server to transmit the scalar value of $g_{\alpha}^{(2)} \approx \thickbar{g}_{[N]}^T\tta_i$, we argue that it is not ideal to directly send this scalar to client $i$, because the calculations of $g_{\alpha}^{(2)}$ and $\nabla_{\alpha_{ij}}F_i(\tta_i, \al_i)$ are during the process of updating $\tta_i$. And since calculating $g_{\alpha}^{(2)}$ involves $\tta_i$, it would be more reasonable to treat $\thickbar{g}_{[N]}^T\tta_i$ as a variable and have it computed by client $i$ locally, rather than treating it as a constant computed it by the server.

In the descriptions of the proposed algorithm above, we did not elaborate on how the client selection procedure would affect the auxiliary risks of each client. As mentioned in \cref{sec:objs_of_pgfed}, in each round $t$, a subset of clients, $\Ss_t$ with $|\Ss_t|=M$, are selected for the training. We, therefore, slightly modify the auxiliary risk from $\mathcal{R}_{aux}^{[N]}$ to $\mathcal{R}_{aux}^{\Ss_t}$. This change will subsequently affect the scope of two terms, namely the auxiliary gradient $\thickTilde{g}_{[N]}$ w.r.t. $\tta_i$ (changed to $\thickTilde{g}_{\Ss_t}$), and $\thickbar{g}_{[N]}$, a portion of the gradient w.r.t. $\alpha_{ij}$ (changed to $\thickbar{g}_{\Ss_t}$). We formally define these two new terms below:
\begin{equation}
    \thickTilde{g}_{\Ss_t}=\mu \sum_{j\in \Ss_t} \alpha_{ij} \nabla_{\tta_j}f_j(\tta_j),
    \label{eq:aux_grad_M}
\end{equation}
\begin{equation}
    \thickbar{g}_{\Ss_t}=\frac{\mu}{M} \sum_{j\in \Ss_t}\nabla_{\tta_j}f_j(\tta_j) .
    \label{eq:grad_alpha_ij_M}
\end{equation}

To sum up, the auxiliary risk $\mathcal{R}_{aux}^{\Ss_t}$ on a client ultimately results in downloading two server-aggregated gradients $\thickTilde{g}_{\Ss_t}$ and $\thickbar{g}_{\Ss_t}$ (and a negligible scalar $g_{\alpha}^{(1)}$). Note that while achieving the explicitness, this server aggregation makes it infeasible for the client to separate other clients' gradient for inferring their data, which protects the clients' privacy.

\subsection{\textbf{\ct{PGFed}} with Momentum (\textbf{\ct{PGFedMo}})}
When considering the client selection for each round, the auxiliary risk changed from $\mathcal{R}_{aux}^{[N]}$ to $\mathcal{R}_{aux}^{\Ss_t}$. This reduces the number of clients' risks involved in the calculation of auxiliary risk from $N$ to $M$, which could be a huge loss of information. To compensate for this loss, each client maintains $\thickTilde{g}_{\Ss_t}^i$ locally, and every time when the client is selected, it updates its local auxiliary gradient $\thickTilde{g}_{\Ss_t}^i$ in a momentum manner, i.e.
\begin{equation}
    \thickTilde{g}_{\Ss_t}^i \leftarrow (1-\beta) \thickTilde{g}_{\Ss_t}^i (\text{downloaded}) + \beta \thickTilde{g}_{\Ss_t}^i (\text{previous})
    \label{eq:momentum}
\end{equation}
In this way, the clients are able to carry the auxiliary gradient, without having to discard the collaborative knowledge from the $N-M$ clients each round. We summarize the proposed algorithm in \cref{alg:PGFed}.

\section{Experiments}
\label{sec:experiments}
\subsection{Experimental Setup}
\label{sec:experimental_setup}


\paragraph{Compared methods.} We evaluate the proposed algorithms, \ct{PGFed} and \ct{PGFedMo}, against a number of state-of-the-art FL algorithms, including two global FL algorithms: the leading \ct{FedAvg} and \ct{FedDyn} that uses a dynamically regularized local training, and nine personalized FL algorithms: \ct{pFedME} which conducts personalization with Moreau Envelopes; \ct{FedFomo} and \ct{APFL} which interpolates personalized models by interpolation; \ct{FedRep}, \ct{LG-FedAvg}, and \ct{FedPer} which personalize different layers of the model; \ct{Per-FedAvg} which leverages meta-learning to learn an initial shared model, and one-step fine-tuning over local data; \ct{FedRoD} which trains a global model with balanced risk and personalized adaptive predictors with empirical risk; and \ct{FedBABU} which focuses on global representation learning and personalize the model by fine-tuning for one epoch.

\paragraph{Datasets.} We conduct experiments on four datasets. In the main content, we report the results on CIFAR10 and CIFAR100~\cite{krizhevsky2009learning}, and we leave the experiments on OrganAMNIST~\cite{medmnistv1} and Office-home~\cite{venkateswara2017deep} dataset in the \textbf{supplementary materials}. Following ~\cite{hsu2019measuring}, we use Dirichlet distribution with $\alpha=0.3$ to partition the dataset into heterogeneous settings with 25, 50, and 100 clients. For each setting, each client's local training and test datasets are under the same distribution. We report the mean top-1 personalized accuracy by averaging the local test accuracies over all clients.


\paragraph{Implementation details.} We adopt the same convolutional neural network (CNN)'s architecture as in \ct{FedAvg} for all the compared methods. There are 2 convolutional layers with 32, and 64 $5\times5$ kernels and 2 fully connected layers with 512 hidden units in this architecture. For the methods that involve representation learning, we split the model into a feature extractor with the first three layers, and a classifier with the last layer. We use a stochastic gradient descent (SGD) optimizer for every method with a fixed momentum of 0.9. Each method is trained on the dataset for 300 federated rounds. For each federated setting, the client sample rate is set to 25\%, and the local training epochs is set to 5. The learning rate is tuned from \{0.1, 0.01, 0.001, 0.0001\} for every method. For \ct{PGFed}, the coefficient for the auxiliary risk, $\mu$, is tuned from \{0.1, 0.05, 0.01, 0.005, 0.001\}. For \ct{PGFedMo}, the momentum, $\beta$, is tuned from \{0.2, 0.5, 0.8\}. More details about hyperparameters values are included in the \textit{Supplementary Material}.

\subsection{Main Results and Analysis}
We report the mean and standard deviation of three different seeds for each setting. As shown in \cref{tab:main_results}, our method achieves the highest mean accuracies under all settings. With large data heterogeneity using Dir(0.3) as in our experiments, local training can often achieve better performance than participating in global FL. As the number of clients increases, the clients benefit more from FL, since local training is more likely to overfit on the small local datasets. \textbf{Within the comparison of personalized FL algorithms, \ct{PGFed} and the momentum upgrade, \ct{PGFedMo}, enjoy performance improvements by up to 15.47\% in accuracy (CIFAR100 with 25 clients)}, outperforming the methods that personalize different components of the model such as \ct{FedRep}, \ct{LG-FedAvg}, and \ct{FedPer}, and even with the fine-tuning in \ct{FedBABU} and the meta-learning \ct{Per-FedAvg}, as well as the interpolation methods such as \ct{APFL} and \ct{FedFomo}. With the global model and the personalized adapter, \ct{FedRod} benefits the local performance by the most within the baselines, yet was still exceeded by \ct{PGFed} and \ct{PGFedMo} by up to $4.63\%$ (on CIFAR100 with 50 clients). For \ct{PGFed}, the momentum upgrade, \ct{PGFedMo}, further boosts its performance since more clients' risk is included in the clients' auxiliary risks with the momentum update.


\begin{table}[ht]
    \centering
    \begin{tabular}{wl{1.6cm}|M{0.4cm}M{0.8cm}|M{0.4cm}M{0.8cm}|M{0.4cm}M{0.8cm}}
    \toprule
    & \multicolumn{2}{c|}{25 clients} & \multicolumn{2}{c|}{50 clients} & \multicolumn{2}{l}{100 clients}\\
    & \scriptsize{\#round} & \scriptsize{speedup} & \scriptsize{\#round} & \scriptsize{speedup} & \scriptsize{\#round} & \scriptsize{speedup}\\\midrule
    \ct{APFL} & $31$ & $1.0\times$ & $28$ & $1.7\times$ & $24$ & $2.6\times$\\
    \ct{FedPer} & $8$ & $3.9\times$ & $6$ & $7.8\times$ & $8$ & $7.9\times$\\
    \footnotesize{\ct{Per-FedAvg}} & $31$ & $1.0\times$ & $47$ & $1.0\times$ & $63$ & $1.0\times$\\
    \ct{FedRoD} & $26$ & $1.2\times$ & $35$ & $1.3\times$ & $10$ & $6.3\times$\\ \midrule
    \ct{PGFed} & $9$ & $3.4\times$ & $14$ & $3.4\times$ & $15$ & $4.2\times$\\
    \ct{PGFedMo} & $9$ & $3.4\times$ & $14$ & $3.4\times$ & $15$ & $4.2\times$\\
    \bottomrule
    \end{tabular}
    \caption{The number of rounds to achieve 70\% mean top-1 personalized accuracy on CIFAR10. The speedup is computed based on the slowest approach listed (i.e. ``1.0$\times$").}
    \label{tab:speedup_cifar10}
\end{table}


In \cref{tab:speedup_cifar10}, we show the convergence speed of the personalized algorithms by reporting the round numbers at which the algorithms achieve 70\% mean top-1 personalized accuracy on CIFAR10. For each setting, we set the algorithm that takes the most round to reach 70\% as ``$1.0\times$'', and find that the proposed \ct{PGFed} and \ct{PGFedMo} consistently present a decent amount of speedup from the compared personalized FL algorithms. Among the six compared personalized FL algorithms, \ct{FedPer} has the fastest convergence speed, due to the smaller size of the globally aggregated model component (only the feature extractor layers are aggregated in \ct{FedPer}). Although not fastest, \ct{PGFed} and \ct{PGFedMo} still have an average of $3.7\times$ speed up in convergence rate, while enjoying the best performance in accuracy.

\begin{table}[ht]
    \centering
    \begin{tabular}{wl{1.5cm}|M{1.7cm}|M{1.7cm}|M{1.7cm}}
    \toprule
    & 25 clients & 50 clients & 100 clients\\ \midrule
    \ct{FedAvg} & \footnotesize{$-8.99\pm10.36$} & \footnotesize{$-8.90\pm15.48$} & \footnotesize{$-5.02\pm14.30$}\\
    \ct{APFL} & $2.79\pm8.07$ & $5.73\pm8.43$ & $8.37\pm6.91$\\
    \ct{FedPer} & $5.31\pm2.56$ & $8.31\pm6.00$ & $8.63\pm5.26$\\
    \footnotesize{\ct{Per-FedAvg}} & $0.72\pm6.22$ & $5.02\pm7.39$ & $8.09\pm7.00$\\
    \ct{FedRoD} & $7.80\pm3.68$ & $8.84\pm6.29$ & $10.68\pm6.14$\\ \midrule
    \ct{PGFed} & $8.49\pm4.67$ & $10.78\pm5.88$ & $11.15\pm5.06$\\
    \ct{PGFedMo} & $8.61\pm3.59$ & $10.90\pm6.11$ & $11.16\pm5.44$\\
    \bottomrule
    \end{tabular}
    \caption{Mean and standard deviation of the individual performance gain over \ct{Local} training in terms of accuracy\% on local test set on CIFAR10.}
    \label{tab:perf_gain_cifar10}
\end{table}

\begin{table}[ht]
    \centering
    \begin{tabular}{wl{1.5cm}|M{1.7cm}|M{1.7cm}|M{1.7cm}}
    \toprule
    & 25 clients & 50 clients & 100 clients\\ \midrule
    \ct{FedAvg} & \footnotesize{$-3.29\pm4.22$} & $0.02\pm4.63$ & $1.77\pm6.38$\\
    \ct{APFL} & $6.48\pm2.93$ & $8.70\pm3.37$ & $9.31\pm4.55$\\
    \ct{FedPer} & $3.43\pm1.80$ & $2.16\pm2.45$ & $2.31\pm3.54$\\
    \footnotesize{\ct{Per-FedAvg}} & $0.07\pm3.71$ & $5.47\pm3.86$ & $7.49\pm5.73$\\
    \ct{FedRoD} & $7.32\pm2.68$ & $6.59\pm3.17$ & $7.47\pm3.69$\\ \midrule
    \ct{PGFed} & $9.34\pm1.71$ & $9.01\pm2.97$ & $12.05\pm3.93$\\
    \ct{PGFedMo} & $9.40\pm1.87$ & $8.99\pm2.76$ & $12.07\pm3.97$\\
    \bottomrule
    \end{tabular}
    \caption{Mean and standard deviation of the individual performance gain over \ct{Local} training in terms of accuracy\% on local test set on CIFAR100.}
    \label{tab:perf_gain_cifar100}
\end{table}

\begin{figure*}[htb]
    \centering
    \includegraphics[width=\linewidth]{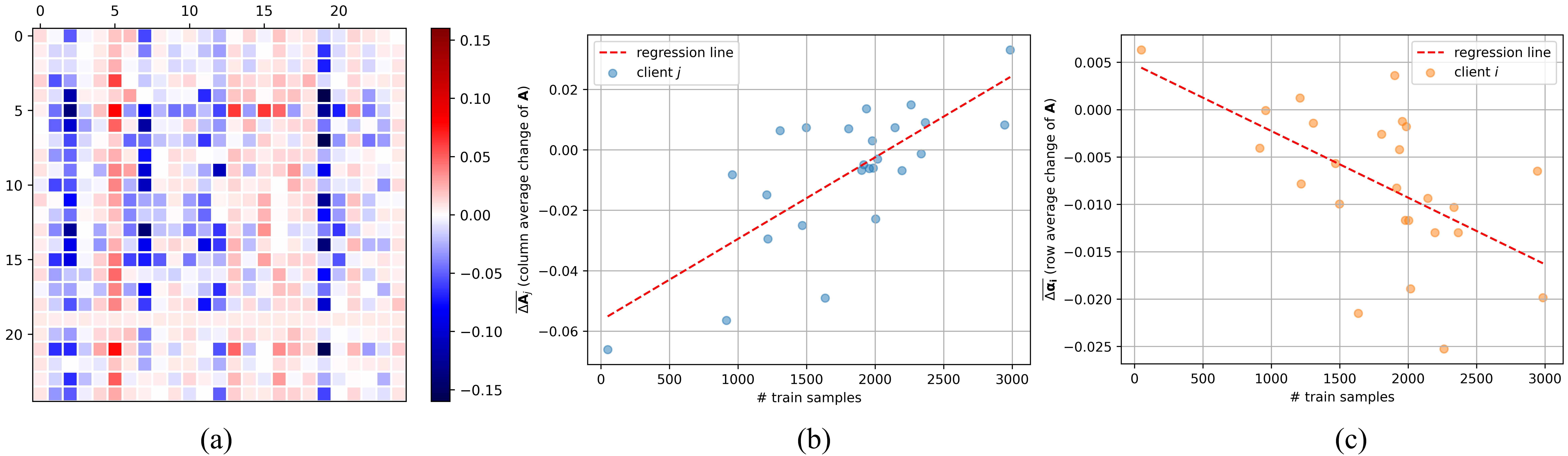}
    \caption{Visualization of the change in $\vb*{A}$. Figure (a) is a heat map of the change in $\vb*{A}$. For Figure (b) and (c), the Y-axis of Figure (b) represents the column average of the change in $\vb*{A}$ (the average change of weights of client $j$'s empirical risk on other clients). The Y-axis of Figure (c) is the row average of the change in $\vb*{A}$ (the average change of weights of the auxiliary risk on client $i$). Through the regression line, we verify the positive correlation between $\overline{\Delta\vb*A_j}$ and $n_j$ in Figure (b), and the negative correlation between $\overline{\Delta\al_i}$ and $n_i$ in Figure (c).}
    \label{fig:alpha}
\end{figure*}

Besides the overall performance and the convergence speed, we take a micro-perspective to examine the individual performance gain over the \ct{Local} training. Specifically, we concentrate on the statistics of the individual performance gain across the federation of clients, which can indicate the fairness performance of the algorithms. For instance, the individual gain with a high standard deviation might indicate that the high performance gains on some clients come with a sacrifice over the performance gains (little or negative gains) on other clients. In \cref{tab:perf_gain_cifar10} and \ref{tab:perf_gain_cifar100}, we show that while enjoying the highest mean individual gains under each setting, the standard deviations of both \ct{PGFed} and \ct{PGFedMo} are reasonably small as well, suggesting a fair boost of performance from every client. We attribute the high mean and low standard deviation merit of the proposed algorithms to the explicitness of our design of local objectives. Although clients' local objectives are personalized, the explicitness enables the clients' local objectives to be of the same form (weighted sum of multiple empirical risks). And each client's local empirical risk takes roughly $1/(1+\mu)$ of its own objective, which is the same for every client, hence the similar individual performance gain and fair federated personalization.

\subsection{Visualization of coefficient matrix $\vb*A$}

We investigate the coefficient matrix, $\vb*{A}$, of $\alpha_{ij}$'s in \ct{PGFed} (recall that $\alpha_{ij}$ is the coefficient of client $j$'s empirical risk on client $i$). Specifically, we visualize the \textit{\textbf{change}} of all $\alpha_{ij} \ \forall i,j\in[N]$ on CIFAR10 with 25 clients. As a refresher, $\al_i$, the $i$-th row of $\vb*{A}$, represents the vector of weights indicating how much client $i$ values other clients' risks. The $j$-th column of $\vb*{A}$, $\vb*{A}_j$, represents the vector of weights indicating how much client $j$ is valuable towards other clients' risks.

In theory, without the first-order estimation (see \cref{eq:pgfed_local_obj}), the gradient of $\alpha_{ij}$ equals $\mu f_j(\tta_i)$, which is non-negative. This suggests that $\alpha_{ij}$ will always converge to 0 given enough time, i.e. $\forall i,j\in [N], \lim_{t \rightarrow \infty} \alpha_{ij}=0$. However, since $f_j(\tta_i)$ is estimated through Taylor expansion, and higher order terms are omitted, in practice of \ct{PGFed}, $\alpha_{ij}$ has a chance to increase (pink and red blocks in \cref{fig:alpha}(a)), or decreases more slowly (light blue in \cref{fig:alpha}(a)), as long as client $i$'s local objective decreases. Therefore, whether $\alpha_{ij}$ changes in the positive or negative directions depends on whether $f_j(\cdot)$ can be ``helpful'' to reduce the local objective of client $i$ as a result of optimization.

We quantify this ``help'' that client $j$ can offer by $\overline{\Delta\vb*A_j}$: the average change of the $j$-th column of $\vb*A$ (i.e. $\alpha_{ij} \ \forall i \in [N]$), and quantify the ``help'' that client $i$ needs by $\overline{\Delta\al_i}$: the average change of the $i$-th row of $\vb*A$ (i.e. $\alpha_{ij} \ \forall j \in [N]$). An intuitive indicator of this ``help'' is the number of local training samples. From the perspective of client $j$ (the helper), larger local training set of client $j$ should be able to make its empirical risk more likely to be helpful to other clients. From the perspective of client $i$ (the helpee), smaller training set of client $i$ should require more help from others. Therefore, ideally, $\overline{\Delta\vb*A_j}$ and $\overline{\Delta\al_i}$ should be positively and negatively correlated to the number of local training samples, respectively. This is verified as a finding in the resulting $\vb*A$ of \ct{PGFed} in \cref{fig:alpha}(b) and \cref{fig:alpha}(c).

\subsection{Generalizability to New Clients}

\begin{figure*}[htb] 
    \includegraphics[width=\linewidth]{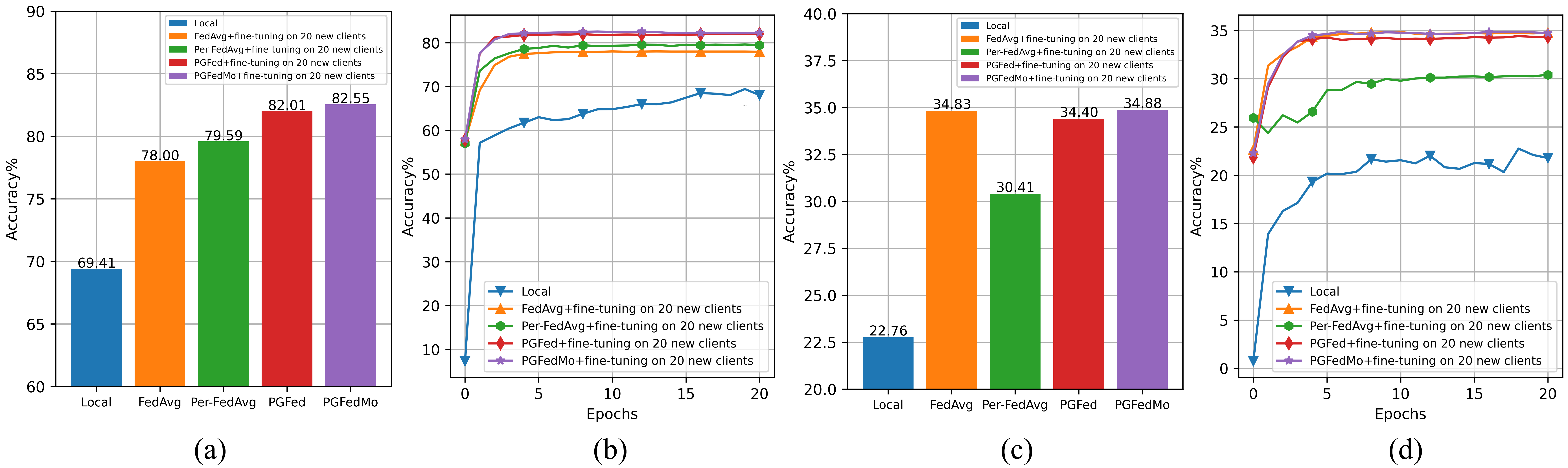}
    \caption{A comparison w.r.t. generalizability on new clients on CIFAR10 and CIFAR100. We fine-tune the global models of different FL approaches and compare them with \ct{Local} training. For FL approaches, we first train the global models on 80 clients for 150 rounds before fine-tuning them on 20 new clients for 20 epochs. Figure (a) and (c) show the final mean personalized accuracies over the 20 new clients on CIFAR10 and CIFAR100, respectively. Figure (b) and (d) show the learning curve along the fine-tuning.}
    \label{fig:80_20}
\end{figure*}

In real-world FL settings, it is possible that some clients did not participate in the FL training, but wish to have a model that could quickly adapt to their local data. This makes it especially desirable for a personalized FL algorithm to possess excellent adaptive ability. By design, \ct{PGFed} is able to generate a global model as a side product (line \hyperref[alg:globmodel]{16} of \cref{alg:PGFed}). Therefore, we simulate such a setting and study the generalizability of our global model, and compare it against several baseline algorithms (global and personalized) that also produce the global model. Specifically, we simulate this setting by conducting FL over 80 clients randomly selected from 100 clients on CIFAR10 and CIFAR100 datasets, and fine-tuning the trained global model on the rest 20 new clients. Besides \ct{Local} training, we compare \ct{PGFed} and \ct{PGFedMo} with \ct{FedAvg} and \ct{Per-FedAvg}, whose goal is to train an easy-to-adapt model. The results are shown in \cref{fig:80_20}.

Since the local fine-tuning on the 20 new clients is a standard vanilla training with SGD for all five compared methods, a high mean personalized accuracy directly indicates stronger overall generalizability of the global model (for FL algorithms), especially when the starting mean accuracies (before fine-tuning) are roughly the same. From \cref{fig:80_20}, we can see that the \ct{Local} trained models do not generalize well on local test data, since the size of the local dataset is likely to be small with a 100-client setting. For the FL algorithms, \ct{PGFedMo} achieves the best mean personalized accuracy, which shows strong generalizability of the global model of the proposed algorithm thanks to the explicit design. While not being the main focus, the high generalizability of \ct{PGFed}'s and \ct{PGFedMo}'s global models shown on new clients also indicates the models' strong adaptiveness in the original personalized FL task. 

\section{Conclusion and Discussion}
\label{sec:conclusion_and_discussion}

In this work, we discovered that a personalized FL algorithm's explicitness enhances models' generalizability, resulting in better local performance. Based on our observations, we proposed, \ct{PGFed}, and its momentum upgrade, \ct{PGFedMo}. Both algorithms explicitly transfer the collaborative knowledge across the clients by formulating their local objectives as personalized global objectives. This is achieved without introducing the seemingly unavoidable massive communication costs or potential privacy risk. Our extensive experiments demonstrated the improvements of the proposed algorithms over state-of-the-art methods on four datasets under different heterogeneous FL settings.


We expect the proposed framework to be extended in different directions. First, since the proposed framework is agnostic, it can be potentially combined with existing implicit FL algorithms such as ~\cite{liang2020think,collins2021exploiting,deng2020adaptive,arivazhagan2019federated,chen2022on,oh2021fedbabu}. Moreover, although the proposed work manages to avoid the $O(N^2)$ communication overhead, since it still costs roughly $2.5\times$ the communication of \ct{FedAvg}, a more communication-efficient method is also worth investigations. We leave these directions for future work.

\section*{Acknowledgements}
This work was supported in part by a National Institutes of Health (NIH) / National Cancer Institute (NCI) grant (1R01CA218405), a National Science Foundation (NSF) grant (CICI: SIVD: 2115082), the grant 1R01EB032896 as part of the NSF/NIH Smart Health and Biomedical Research in the Era of Artificial Intelligence and Advanced Data Science Program, a NIH R01 Supplement grant (3R01EB032896-03S1), a Pitt Momentum Funds scaling award (Pittsburgh Center for AI Innovation in Medical Imaging), and an Amazon AWS Machine Learning Research Award. This work used the Extreme Science and Engineering Discovery Environment (XSEDE), which is supported by NSF grant number ACI-1548562. Specifically, it used the Bridges-2 system, which is supported by NSF award number ACI-1928147, at the Pittsburgh Supercomputing Center. This work was also partially supported by the NSF/Intel Partnership on MLWiNS under Grant No. 2003198.

{\small
\bibliographystyle{ieee_fullname}
\bibliography{egbib}

\begin{thebibliography}{10}\itemsep=-1pt

\bibitem{acar2020federated}
Durmus Alp~Emre Acar, Yue Zhao, Ramon Matas, Matthew Mattina, Paul Whatmough,
  and Venkatesh Saligrama.
\newblock Federated learning based on dynamic regularization.
\newblock In {\em International Conference on Learning Representations}, 2020.

\bibitem{arivazhagan2019federated}
Manoj~Ghuhan Arivazhagan, Vinay Aggarwal, Aaditya~Kumar Singh, and Sunav
  Choudhary.
\newblock Federated learning with personalization layers.
\newblock {\em arXiv preprint arXiv:1912.00818}, 2019.

\bibitem{chen2018federated}
Fei Chen, Mi Luo, Zhenhua Dong, Zhenguo Li, and Xiuqiang He.
\newblock Federated meta-learning with fast convergence and efficient
  communication.
\newblock {\em arXiv preprint arXiv:1802.07876}, 2018.

\bibitem{chen2022on}
Hong-You Chen and Wei-Lun Chao.
\newblock On bridging generic and personalized federated learning for image
  classification.
\newblock In {\em International Conference on Learning Representations}, 2022.

\bibitem{collins2021exploiting}
Liam Collins, Hamed Hassani, Aryan Mokhtari, and Sanjay Shakkottai.
\newblock Exploiting shared representations for personalized federated
  learning.
\newblock In {\em International Conference on Machine Learning}, pages
  2089--2099. PMLR, 2021.

\bibitem{deng2020adaptive}
Yuyang Deng, Mohammad~Mahdi Kamani, and Mehrdad Mahdavi.
\newblock Adaptive personalized federated learning.
\newblock {\em arXiv preprint arXiv:2003.13461}, 2020.

\bibitem{fallah2020personalized}
Alireza Fallah, Aryan Mokhtari, and Asuman Ozdaglar.
\newblock Personalized federated learning with theoretical guarantees: A
  model-agnostic meta-learning approach.
\newblock {\em Advances in Neural Information Processing Systems},
  33:3557--3568, 2020.

\bibitem{finn2017model}
Chelsea Finn, Pieter Abbeel, and Sergey Levine.
\newblock Model-agnostic meta-learning for fast adaptation of deep networks.
\newblock In {\em International Conference on Machine Learning}, pages
  1126--1135. PMLR, 2017.

\bibitem{ghosh2020efficient}
Avishek Ghosh, Jichan Chung, Dong Yin, and Kannan Ramchandran.
\newblock An efficient framework for clustered federated learning.
\newblock {\em Advances in Neural Information Processing Systems},
  33:19586--19597, 2020.

\bibitem{hospedales2021meta}
Timothy Hospedales, Antreas Antoniou, Paul Micaelli, and Amos Storkey.
\newblock Meta-learning in neural networks: A survey.
\newblock {\em IEEE transactions on pattern analysis and machine intelligence},
  44(9):5149--5169, 2021.

\bibitem{hsu2019measuring}
Tzu-Ming~Harry Hsu, Hang Qi, and Matthew Brown.
\newblock Measuring the effects of non-identical data distribution for
  federated visual classification.
\newblock {\em arXiv preprint arXiv:1909.06335}, 2019.

\bibitem{huang2021personalized}
Yutao Huang, Lingyang Chu, Zirui Zhou, Lanjun Wang, Jiangchuan Liu, Jian Pei,
  and Yong Zhang.
\newblock Personalized cross-silo federated learning on non-iid data.
\newblock In {\em Proceedings of the AAAI Conference on Artificial
  Intelligence}, volume~35, pages 7865--7873, 2021.

\bibitem{jiang2019improving}
Yihan Jiang, Jakub Kone{\v{c}}n{\`y}, Keith Rush, and Sreeram Kannan.
\newblock Improving federated learning personalization via model agnostic meta
  learning.
\newblock {\em arXiv preprint arXiv:1909.12488}, 2019.

\bibitem{kairouz2019advances}
Peter Kairouz, H~Brendan McMahan, Brendan Avent, Aur{\'e}lien Bellet, Mehdi
  Bennis, Arjun~Nitin Bhagoji, Kallista Bonawitz, Zachary Charles, Graham
  Cormode, Rachel Cummings, et~al.
\newblock Advances and open problems in federated learning.
\newblock {\em arXiv preprint arXiv:1912.04977}, 2019.

\bibitem{karimireddy2020scaffold}
Sai~Praneeth Karimireddy, Satyen Kale, Mehryar Mohri, Sashank Reddi, Sebastian
  Stich, and Ananda~Theertha Suresh.
\newblock Scaffold: Stochastic controlled averaging for federated learning.
\newblock In {\em International Conference on Machine Learning}, pages
  5132--5143. PMLR, 2020.

\bibitem{khan2021federated}
Latif~U Khan, Walid Saad, Zhu Han, Ekram Hossain, and Choong~Seon Hong.
\newblock Federated learning for internet of things: Recent advances, taxonomy,
  and open challenges.
\newblock {\em IEEE Communications Surveys \& Tutorials}, 2021.

\bibitem{krizhevsky2009learning}
Alex Krizhevsky, Geoffrey Hinton, et~al.
\newblock Learning multiple layers of features from tiny images.
\newblock 2009.

\bibitem{kulkarni2020survey}
Viraj Kulkarni, Milind Kulkarni, and Aniruddha Pant.
\newblock Survey of personalization techniques for federated learning.
\newblock In {\em 2020 Fourth World Conference on Smart Trends in Systems,
  Security and Sustainability (WorldS4)}, pages 794--797. IEEE, 2020.

\bibitem{li2019fedmd}
Daliang Li and Junpu Wang.
\newblock Fedmd: Heterogenous federated learning via model distillation.
\newblock {\em arXiv preprint arXiv:1910.03581}, 2019.

\bibitem{li2021model}
Qinbin Li, Bingsheng He, and Dawn Song.
\newblock Model-contrastive federated learning.
\newblock In {\em Proceedings of the IEEE/CVF Conference on Computer Vision and
  Pattern Recognition}, pages 10713--10722, 2021.

\bibitem{li2020federateda}
Tian Li, Anit~Kumar Sahu, Ameet Talwalkar, and Virginia Smith.
\newblock Federated learning: Challenges, methods, and future directions.
\newblock {\em IEEE Signal Processing Magazine}, 37(3):50--60, 2020.

\bibitem{li2020federated}
Tian Li, Anit~Kumar Sahu, Manzil Zaheer, Maziar Sanjabi, Ameet Talwalkar, and
  Virginia Smith.
\newblock Federated optimization in heterogeneous networks.
\newblock {\em Proceedings of Machine Learning and Systems}, 2:429--450, 2020.

\bibitem{liang2020think}
Paul~Pu Liang, Terrance Liu, Liu Ziyin, Nicholas~B Allen, Randy~P Auerbach,
  David Brent, Ruslan Salakhutdinov, and Louis-Philippe Morency.
\newblock Think locally, act globally: Federated learning with local and global
  representations.
\newblock {\em arXiv preprint arXiv:2001.01523}, 2020.

\bibitem{long2020federated}
Guodong Long, Yue Tan, Jing Jiang, and Chengqi Zhang.
\newblock Federated learning for open banking.
\newblock In {\em Federated learning}, pages 240--254. Springer, 2020.

\bibitem{luo2021adapt}
Jun Luo and Shandong Wu.
\newblock Adapt to adaptation: Learning personalization for cross-silo
  federated learning.
\newblock {\em arXiv preprint arXiv:2110.08394}, 2021.

\bibitem{mcmahan2017communication}
Brendan McMahan, Eider Moore, Daniel Ramage, Seth Hampson, and Blaise~Aguera y
  Arcas.
\newblock Communication-efficient learning of deep networks from decentralized
  data.
\newblock In {\em Artificial intelligence and statistics}, pages 1273--1282.
  PMLR, 2017.

\bibitem{mendieta2022local}
Matias Mendieta, Taojiannan Yang, Pu Wang, Minwoo Lee, Zhengming Ding, and Chen
  Chen.
\newblock Local learning matters: Rethinking data heterogeneity in federated
  learning.
\newblock In {\em Proceedings of the IEEE/CVF Conference on Computer Vision and
  Pattern Recognition}, pages 8397--8406, 2022.

\bibitem{oh2021fedbabu}
Jaehoon Oh, Sangmook Kim, and Se-Young Yun.
\newblock Fedbabu: Towards enhanced representation for federated image
  classification.
\newblock {\em arXiv preprint arXiv:2106.06042}, 2021.

\bibitem{pillutla2022federated}
Krishna Pillutla, Kshitiz Malik, Abdel-Rahman Mohamed, Mike Rabbat, Maziar
  Sanjabi, and Lin Xiao.
\newblock Federated learning with partial model personalization.
\newblock In {\em International Conference on Machine Learning}, pages
  17716--17758. PMLR, 2022.

\bibitem{sahu2018convergence}
Anit~Kumar Sahu, Tian Li, Maziar Sanjabi, Manzil Zaheer, Ameet Talwalkar, and
  Virginia Smith.
\newblock On the convergence of federated optimization in heterogeneous
  networks.
\newblock {\em arXiv preprint arXiv:1812.06127}, 3:3, 2018.

\bibitem{sattler2020byzantine}
Felix Sattler, Klaus-Robert M{\"u}ller, Thomas Wiegand, and Wojciech Samek.
\newblock On the byzantine robustness of clustered federated learning.
\newblock In {\em ICASSP 2020-2020 IEEE International Conference on Acoustics,
  Speech and Signal Processing (ICASSP)}, pages 8861--8865. IEEE, 2020.

\bibitem{sattler2019robust}
Felix Sattler, Simon Wiedemann, Klaus-Robert M{\"u}ller, and Wojciech Samek.
\newblock Robust and communication-efficient federated learning from non-iid
  data.
\newblock {\em IEEE transactions on neural networks and learning systems},
  31(9):3400--3413, 2019.

\bibitem{shingi2020federated}
Geet Shingi.
\newblock A federated learning based approach for loan defaults prediction.
\newblock In {\em 2020 International Conference on Data Mining Workshops
  (ICDMW)}, pages 362--368. IEEE, 2020.

\bibitem{smith2017federated}
Virginia Smith, Chao-Kai Chiang, Maziar Sanjabi, and Ameet Talwalkar.
\newblock Federated multi-task learning.
\newblock {\em arXiv preprint arXiv:1705.10467}, 2017.

\bibitem{t2020personalized}
Canh T~Dinh, Nguyen Tran, and Josh Nguyen.
\newblock Personalized federated learning with moreau envelopes.
\newblock {\em Advances in Neural Information Processing Systems},
  33:21394--21405, 2020.

\bibitem{tan2022towards}
Alysa~Ziying Tan, Han Yu, Lizhen Cui, and Qiang Yang.
\newblock Towards personalized federated learning.
\newblock {\em IEEE Transactions on Neural Networks and Learning Systems},
  2022.

\bibitem{venkateswara2017deep}
Hemanth Venkateswara, Jose Eusebio, Shayok Chakraborty, and Sethuraman
  Panchanathan.
\newblock Deep hashing network for unsupervised domain adaptation.
\newblock In {\em Proceedings of the IEEE Conference on Computer Vision and
  Pattern Recognition}, pages 5018--5027, 2017.

\bibitem{voigt2017eu}
Paul Voigt and Axel Von~dem Bussche.
\newblock The eu general data protection regulation (gdpr).
\newblock {\em A Practical Guide, 1st Ed., Cham: Springer International
  Publishing}, 10(3152676):10--5555, 2017.

\bibitem{wang2021field}
Jianyu Wang, Zachary Charles, Zheng Xu, Gauri Joshi, H~Brendan McMahan, Maruan
  Al-Shedivat, Galen Andrew, Salman Avestimehr, Katharine Daly, Deepesh Data,
  et~al.
\newblock A field guide to federated optimization.
\newblock {\em arXiv preprint arXiv:2107.06917}, 2021.

\bibitem{wang2019federated}
Kangkang Wang, Rajiv Mathews, Chlo{\'e} Kiddon, Hubert Eichner, Fran{\c{c}}oise
  Beaufays, and Daniel Ramage.
\newblock Federated evaluation of on-device personalization.
\newblock {\em arXiv preprint arXiv:1910.10252}, 2019.

\bibitem{xu2021federated}
Jie Xu, Benjamin~S Glicksberg, Chang Su, Peter Walker, Jiang Bian, and Fei
  Wang.
\newblock Federated learning for healthcare informatics.
\newblock {\em Journal of Healthcare Informatics Research}, 5(1):1--19, 2021.

\bibitem{medmnistv1}
Jiancheng Yang, Rui Shi, and Bingbing Ni.
\newblock Medmnist classification decathlon: A lightweight automl benchmark for
  medical image analysis.
\newblock In {\em IEEE 18th International Symposium on Biomedical Imaging
  (ISBI)}, pages 191--195, 2021.

\bibitem{zhang2020personalized}
Michael Zhang, Karan Sapra, Sanja Fidler, Serena Yeung, and Jose~M Alvarez.
\newblock Personalized federated learning with first order model optimization.
\newblock {\em arXiv preprint arXiv:2012.08565}, 2020.

\bibitem{zhang2017survey}
Yu Zhang and Qiang Yang.
\newblock A survey on multi-task learning.
\newblock {\em arXiv preprint arXiv:1707.08114}, 2017.

\end{thebibliography}
}

\clearpage
\pagebreak

\appendix
\addcontentsline{toc}{section}{Appendices}
\section{Overview}
\label{sec:overview}
We organize the supplementary material as follows:
\begin{itemize}
    \item In \cref{sec:additional_experiments_and_analyses}, we report additional results and analyses of \ct{PGFed} and conduct experiments on two other datasets, OrganAMNIST~\cite{medmnistv1} \& Office-home~\cite{venkateswara2017deep} with different FL settings.
    \item In \cref{sec:computation}, we compare the local computational speed of the proposed algorithm, \ct{PGFed}, with the baselines that achieved top performance in the experiments on CIFAR10 and CIFAR100.
    \item In \cref{sec:pgfedre}, we further propose \ct{PGFed-CE}, a variation on top of \ct{PGFed} to reduce both the communication and the computation cost simultaneously.
    \item In \cref{sec:hyperparameters}, we report details in hyperparameters regarding our experiments.
\end{itemize}

\section{Additional Experiments and Analyses}
\label{sec:additional_experiments_and_analyses}
\subsection{Convergence Behavior}


\begin{figure*}[htb]
    \centering
    \includegraphics[width=\linewidth]{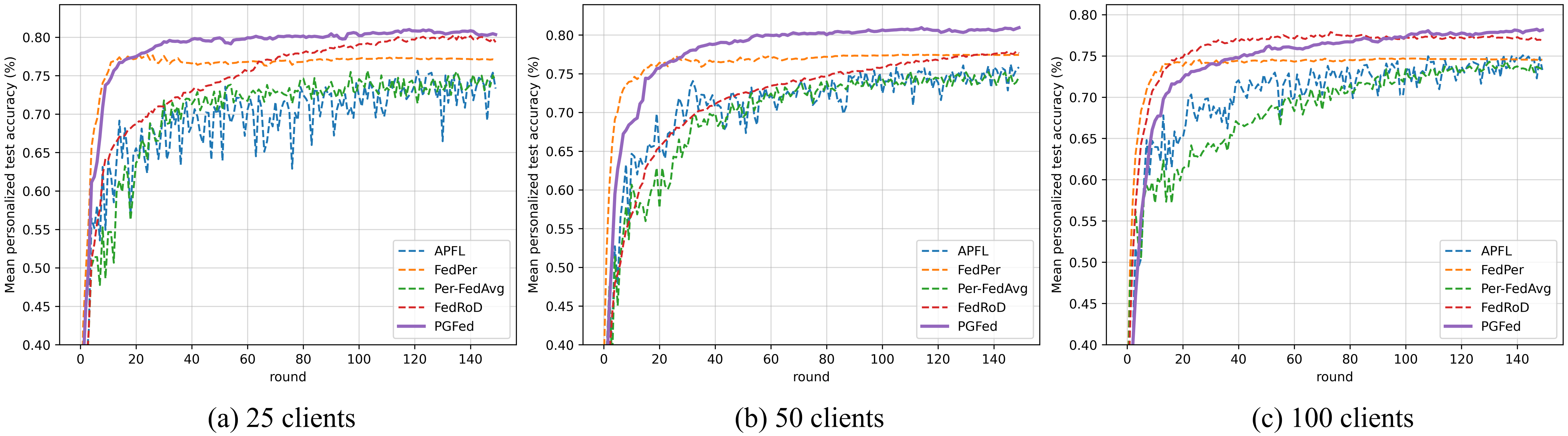}
    \caption{Convergence behavior of the personalized FL approaches with top performance on CIFAR10. While achieving the highest accuracy performance, \ct{PGFed} is also able to consistently converge faster than several of the baselines that reach high accuracies.}
    \label{fig:converg}
\end{figure*}

We empirically study the convergence behavior of \ct{PGFed} and the baselines that achieved high performance on CIFAR10 and CIFAR100. For each method, we plot its mean personalized test accuracy on CIFAR10 for the first 150 rounds of training under 25-, 50- and 100-client settings, as shown in \cref{fig:converg}.

From the results, we can see that, while achieving the highest accuracy performance, \ct{PGFed} is also able to consistently converge faster than most of the baselines that reach high accuracies. Fast as it is under these settings, one limitation of \ct{PGFed} is that the update of $\alpha_{ij}$ in \ct{PGFed} only happens when client $i$ and client $j$ are selected in two consecutive rounds (client $j$ is selected exactly one round before client $i$), which happens by chance. Therefore, this randomness might slightly limit the overall convergence behavior of \ct{PGFed}, but it is the existence of $\alpha_{ij}$'s that enables the adaptive personalization of how much each client values other clients' empirical risks, hence the higher accuracy.

\subsection{Experiments on Other Datasets}
To further evaluate the effectiveness of \ct{PGFed} with different types of data and different FL settings, we conduct experiments on two more datasets: OrganAMNIST~\cite{medmnistv1} and Office-home~\cite{venkateswara2017deep}. OrganAMNIST is a medical imaging dataset of abdominal CT images with 11 classes. Office-home~\cite{venkateswara2017deep} contains four domains (Art, Clipart, Product, and Real World) of images depicting 65 classes of objects typically found in Office and Home settings.

\begin{table}[ht]
    \centering
    \begin{tabular}{wl{1.5cm}|M{1.7cm}|M{1.7cm}|M{1.7cm}}
    \toprule
    & 25 clients\footnotesize{\newline{sample 50\%} Dir(1.0)} & 50 clients\footnotesize{\newline{sample 25\%} Dir(0.3)} & 100 clients\footnotesize{\newline{sample 25\%} Dir(0.3)}\\ \midrule
    \ct{Local} & $90.45\pm0.19$ & $90.63\pm0.07$ & $87.14\pm0.10$ \\
    \ct{FedAvg} & $99.11\pm0.03$ & $98.74\pm0.04$ & $98.47\pm0.08$ \\
    \ct{APFL} & $97.49\pm0.05$ & $97.53\pm0.06$ & $96.19\pm0.11$ \\
    \ct{FedRep} & $95.06\pm0.16$ & $94.86\pm0.07$ & $92.47\pm0.04$ \\
    \ct{LGFedAvg} & $90.47\pm0.18$ & $90.99\pm0.08$ & $87.52\pm0.22$ \\
    \ct{FedPer} & $97.89\pm0.06$ & $97.55\pm0.08$ & $95.56\pm0.33$ \\
    \footnotesize{\ct{Per-FedAvg}} & $98.40\pm0.02$ & $96.80\pm0.04$ & $95.09\pm0.07$ \\
    \ct{FedRoD} & $98.61\pm0.05$ & $98.14\pm0.09$ & $97.05\pm0.06$ \\
    \ct{FedBABU} & $96.49\pm0.28$ & $94.33\pm0.13$ & $91.07\pm0.23$ \\ \midrule
    \ct{PGFed} & $99.20\pm0.04$ & $\vb{99.17}$$\pm$$\vb{0.05}$ & $\vb{98.94}$$\pm$$\vb{0.02}$ \\
    \ct{PGFedMo} & $\vb{99.21}$$\pm$$\vb{0.04}$ & $99.17\pm0.07$ & $98.86\pm0.06$ \\
    \bottomrule
    \end{tabular}
    \caption{Mean and standard deviation over three trials of the mean personalized test accuracy (\%) on OrganAMNIST}
    \label{tab:medmnist_perf}
\end{table}

\begin{table*}[ht]
    \centering
    \begin{tabular}{wl{2.5cm}|M{2.4cm}M{2.4cm}M{2.4cm}M{2.4cm}|M{2.4cm}}
    \toprule
    & Art & Clipart & Product & Real World & Mean\\ \midrule
    \ct{Local} & $17.16\pm0.85$ & $37.65\pm0.47$ & $43.83\pm0.40$ & $24.50\pm0.21$ & $30.79\pm0.23$ \\
    \ct{FedAvg} & $11.68\pm1.26$ & $41.29\pm0.85$ & $42.49\pm1.28$ & $19.14\pm0.89$ & $28.65\pm0.49$ \\
    \ct{APFL} & $19.11\pm1.55$ & $44.67\pm0.61$ & $\vb{50.40\pm0.56}$ & $25.85\pm0.88$ & $35.00\pm0.41$ \\
    \ct{FedRep} & $20.24\pm1.45$ & $38.43\pm1.02$ & $43.70\pm1.04$ & $24.02\pm0.81$ & $31.60\pm0.05$ \\
    \ct{LGFedAvg} & $17.54\pm0.45$ & $38.75\pm0.13$ & $44.59\pm0.62$ & $25.79\pm0.61$ & $31.67\pm0.21$ \\
    \ct{FedPer} & $17.83\pm1.07$ & $38.97\pm0.35$ & $45.87\pm0.13$ & $25.01\pm0.52$ & $31.92\pm0.24$ \\
    \ct{Per-FedAvg} & $14.62\pm0.40$ & $39.94\pm1.29$ & $44.40\pm1.32$ & $21.58\pm0.65$ & $30.13\pm0.07$ \\
    \ct{FedRoD} & $19.67\pm1.23$ & $42.44\pm0.77$ & $44.34\pm2.07$ & $24.28\pm1.69$ & $32.68\pm0.69$ \\
    \ct{FedBABU} & $18.18\pm3.54$ & $42.10\pm2.31$ & $43.51\pm0.91$ & $\vb{26.81\pm1.86}$ & $33.38\pm0.29$ \\ \midrule
    \ct{PGFed} & $\vb{22.40\pm0.26}$ & $\vb{46.48\pm1.00}$ & \underline{$49.86\pm2.14$} & $26.04\pm0.80$ & $\vb{36.19\pm0.92}$ \\
    \ct{PGFedMo} & \underline{$22.16\pm0.45$} & \underline{$45.88\pm0.83$} & $49.45\pm0.19$ & \underline{$26.60\pm0.99$} & \underline{$36.02\pm0.20$} \\
    \bottomrule
    \end{tabular}
    \caption{Mean and standard deviation over three trials of the mean personalized accuracy\% of the four domains (5 clients/domain) and the average performance on Office-home dataset. The highest and second-highest accuracies under each setting are in \textbf{bold} and \underline{underlined}, respectively.}
    \label{tab:office-home-20clients}
\end{table*}


For OrganAMNIST, we adopt three settings with different numbers (25, 50, 100) of clients. For the 50- and 100-client settings, we follow the same setting as in the experiments on CIFAR10/CIFAR100, and use the Dirichlet distribution with $\alpha=0.3$ (Dir(0.3)) and 25\% client sample rate for each round. For the 25-client setting, we reduce the heterogeneity in the dataset via Dir(1.0) distribution, and use a higher client sample rate (50\%) to simulate a situation more similar to cross-silo FL settings. For Office-home, we adopt a 20-client setting, where each domain contains 5 clients. The non-IIDness in each domain is achieved by Dir(0.3). The mean personalized test accuracies of each domain and over the whole federation are reported. We compare the proposed \ct{PGFed} and \ct{PGFedMo} against \ct{Local}, \ct{FedAvg}, and the personalized FL baselines that achieved high performance in previous experiments on CIFAR10 and CIFAR100. The results are shown in \cref{tab:medmnist_perf} and \cref{tab:office-home-20clients}.

\begin{figure}[t]
  \centering
   \includegraphics[width=0.9\linewidth]{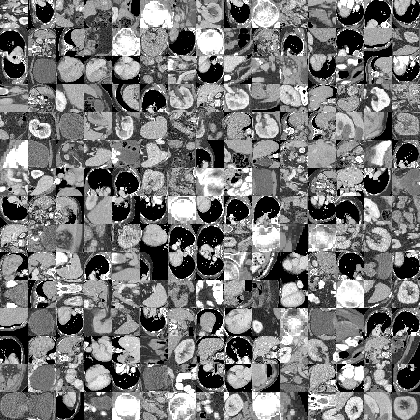}
   \caption{OrganAMNIST~\cite{venkateswara2017deep} image samples.}
   \label{fig:organamnist}
\end{figure}


For OrganAMNIST, \ct{PGFed} and \ct{PGFedMo} achieve the best performance under all three settings. In addition, the proposed algorithms do not have an obvious drop in the performance from the less heterogeneous 25-client setting to 50-client and 100-client settings. This is not the case for many other personalized FL baselines (\ct{FedPer}, \ct{Per-FedAvg}, \ct{FedRoD}, and \ct{FedBABU}). Moreover, \ct{FedAvg} achieves excellent performance on OrganAMNIST due to the similarity of clients' images (see \cref{fig:organamnist}). Since the Dirichlet distribution can only differ the clients in $P(y)$, the label distribution, instead of $P(x|y)$, the distribution of the images given the label, a simple averaging might work just fine on OrganAMNIST compared with other datasets. On the contrary, Office-home is a dataset that addresses different $P(x|y)$ since for this dataset, even the images within the same class can be dramatically different if they belong to clients from different domains, which is indicated by the worse performance of \ct{FedAvg} than \ct{Local} training. Results on this dataset also show that \ct{PGFed} and \ct{PGFedMo} consistently outperform most of the compared methods in each domain, and achieve the highest mean accuracies over all domains, demonstrating their superiority over all the compared methods.

\section{Comparison in Local Computational Speed}
\label{sec:computation}

In this section, we study the local computational speed of \ct{PGFed} and the baselines that achieved top performance in the experiments. We measure the local computational speed by the number of images each method can process per second, and report the local computational speed of the methods in \cref{tab:computational_speed} on CIFAR10 with 50 clients and a batch size of 128, using an NVIDIA Tesla V100 GPU and an Intel(R) Xeon(R) Gold 6248 CPU.

\begin{table}[ht]
    \centering
    \begin{threeparttable}
    \begin{tabular}{wl{1.8cm}|M{1.4cm}|M{1.7cm}|M{1.7cm}}
    \toprule
    & Images/s & \footnotesize{Relative speed} & Accuracy\\ \midrule
    \ct{FedAvg} & $6917.1$ & $100.00\%$ & $64.41\pm0.66$ \\
    \ct{APFL} & $3389.8$ & $48.99\%$ & $77.36\pm0.18$ \\
    \footnotesize{\ct{Per-FedAvg}} & $3464.5$ & $50.09\%$ & $76.27\pm0.50$ \\
    \ct{FedRoD} & $6682.4$ & $96.61\%$ & $79.61\pm0.22$ \\ \midrule
    \ct{PGFed} & $6120.0$ & $88.48\%$ & $81.42\pm0.31$ \\
    \ct{PGFedMo} & $6032.8$ & $87.22\%$ & $81.48\pm0.32$ \\
    \ct{PGFed-CE}\tnote{*}  & $6175.5$ & $89.28\%$ & $81.16\pm0.56$ \\
    \bottomrule
    \end{tabular}
    \begin{tablenotes}
        \item[*]{\scriptsize{A more communication-efficient variation of \ct{PGFed}, introduced in \cref{sec:pgfedre}}}
    \end{tablenotes}
    \end{threeparttable}
    \caption{Computational speed (in terms of ``images/s'') and accuracy on CIFAR10 with 50 clients}
    \label{tab:computational_speed}
\end{table}

From the results, we can see that \ct{PGFed} not only reaches high accuracy, but has a relatively high computational speed as well. With a batch size of 128, \ct{PGFed} reaches a speed equivalent to 88.48\% of \ct{FedAvg}'s speeds. \ct{PGFedMo} is slightly slower than \ct{PGFed} due to the momentum update of the auxiliary gradient. However, for some of the compared methods that also achieve high accuracy, their computational speed is compromised by around 50\% (compared to \ct{FedAvg}): \ct{APFL} needs to train a global model and a local adapter, while \ct{Per-FedAvg} leverages meta-learning which is a bi-level optimization problem. These methods either train two models or conduct twice gradient descent for each iteration. \ct{FedRoD} trains a global model and a local classifier, which ends up being 8.13\% faster than \ct{PGFed}. For \ct{PGFed}, the extra local computation (over \ct{FedAvg}) happens at the addition of the gradients from both the local empirical risk and the auxiliary risk, and at the update of $\al_i$ where a dot product of vectorized models is calculated (see Eq.(12) in the main paper).

\section{\ct{PGFed-CE}, a More Communication- and Computation-efficient \ct{PGFed}}
\label{sec:pgfedre}
As mentioned in Sec. 6 of the main paper, although \ct{PGFed} manages to circumvent the seemingly unavoidable $O(N^2)$ communication cost, and achieve asymptotically the same communication cost as \ct{FedAvg} ($O(N)$), since each client is required to download three and upload two models/gradients per round, on average the communication cost is still high (roughly 2.5 times as much as that of \ct{FedAvg}).

In this section, we provide a more communication-efficient version of \ct{PGFed}, dubbed \ct{PGFed-CE} that downloads one less gradient from the server. In Sec. 4 of the main paper, we mentioned that $g_{\alpha}^{(2)}$, a portion of the gradient of the local objective in terms of $\al_i$ in \ct{PGFed}, could be computed on the server instead of the client. This is because
\begin{equation}
    g_{\alpha}^{(2)} = \mu \nabla_{\tta_j}f_j(\tta_j)^T\tta_i,
\end{equation}
and the server has both $\nabla_{\tta_j}f_j(\tta_j)$ from the previous round and the current round global model as the initialization of $\tta_i$. Therefore, it is possible to treat $g_{\alpha}^{(2)}=\mu \nabla_{\tta_j}f_j(\tta_j)^T \tta_{glob}$ as a constant computed by the server, where the global model is used as an estimation of $\tta_i$ for the whole round. Since $\alpha_i$ should change adaptively according to the change of $\tta_i$ during local training, using the global model as a fixed estimation is not ideal. Nonetheless, this variation saves the communication cost by the size of one gradient ($\thickbar{g}_{\Ss_t}$) from clients' round-beginning download, that \textit{was} needed to adaptively compute $\al_i$ locally, which also slightly saves the local computation.

We name this more communication- and computation-efficient variant of \ct{PGFed} as \ct{PGFed-CE}. In \ct{PGFed-CE}, each client is now required to download two, instead of three, models/gradients per round. Consequently, on average, the communication cost is reduced from 2.5 to 2 times as much as that of \ct{FedAvg}. In addition, we follow \cref{sec:computation} and report the local computational speed of \ct{PGFed-CE} and its performance on CIFAR10 with 50 clients in \cref{tab:computational_speed}. As expected, besides the reduced communication cost, \ct{PGFed-CE} also simultaneously increases the local computational speed with little drop in the accuracy.

\section{Hyperparameters}
\label{sec:hyperparameters}
Besides the hyperparameter tuning of \ct{PGFed} reported in Sec. 4.1 in the main paper, we further report the hyperparameter tuning of the compared baselines in this section. The learning rate of all baselines is tuned from \{0.1, 0.01, 0.001, 0.0001\}. For \ct{FedDyn}, we tuned the $\alpha$ in \{0.1, 0.01, 0.001\}. For \ct{APFL}, the $\alpha$ is tuned in \{0.25, 0.50, 0.75\}. For \ct{Per-FedAvg}, the two learning rates for each step are selected from \{0.1, 0.01, 0.001, 0.0001\}, which is the same for \ct{FedBABU}'s learning rates for federated training and fine-tuning step.

\end{document}